\documentclass[letterpaper]{article} 

\ifdefined\aaaianonymous
    \usepackage[submission]{aaai2026}  
\else
    \usepackage{aaai2026}              
\fi

\nocopyright 
\usepackage{pdfpages}
\usepackage{amsmath}
\usepackage{tcolorbox}
\usepackage{textcomp}
\usepackage{enumitem}
\usepackage{hyperref}
\usepackage{times}  
\usepackage{helvet}  
\usepackage{courier}  
\usepackage{graphicx} 
\usepackage{natbib}  
\usepackage{caption} 
\usepackage{algorithm}
\usepackage{algorithmic}

\usepackage{newfloat}
\usepackage{listings}
\DeclareCaptionStyle{ruled}{labelfont=normalfont,labelsep=colon,strut=off} 
\floatstyle{ruled}
\newfloat{listing}{tb}{lst}{}
\floatname{listing}{Listing}

\usepackage{graphicx} 
\usepackage{amssymb}  

\usepackage{booktabs}
\usepackage{multirow} 
\usepackage{makecell} 

\usepackage{adjustbox}

\usepackage{pifont}      
\usepackage{subcaption}  
\usepackage{comment}     
\usepackage{fontawesome}

\usepackage{cleveref}       
\usepackage{color}

\usepackage{arydshln}

\ifdefined\aaaianonymous
    \title{ViDA-UGC: Detailed Image Quality Analysis via Visual Distortion Assessment for UGC Images}
\else
    \title{ViDA-UGC: Detailed Image Quality Analysis via Visual Distortion Assessment for UGC Images}
\fi
\author{
    Wenjie Liao\textsuperscript{\rm 1,2}\thanks{Intern in Media Evaluation Lab, ByteDance Inc.},  
    Jieyu Yuan\textsuperscript{\rm 1},
    Yifang Xu\textsuperscript{\rm 2},
    Chunle Guo\textsuperscript{\rm 1},
    Zilong Zhang\textsuperscript{\rm 1},\\
    Jihong Li\textsuperscript{\rm 1},
    Jiachen Fu\textsuperscript{\rm 1},
    Haotian Fan\textsuperscript{\rm 2}\textsuperscript{†},  
    Tao Li\textsuperscript{\rm 2},
    Junhui Cui\textsuperscript{\rm 2},
    Chongyi Li\textsuperscript{\rm 1}\textsuperscript{‡}  
}
\affiliations{
    \textsuperscript{\rm 1}VCIP, CS, Nankai University\\
    \textsuperscript{\rm 2}Bytedance Inc.\\
    \textsuperscript{†}: Project lead \quad  
    \textsuperscript{‡}: Corresponding author  
}

\usepackage{bibentry}

\begin{document}

\maketitle

\begin{figure*}[!t]
\centering
\includegraphics[width=0.9\textwidth]{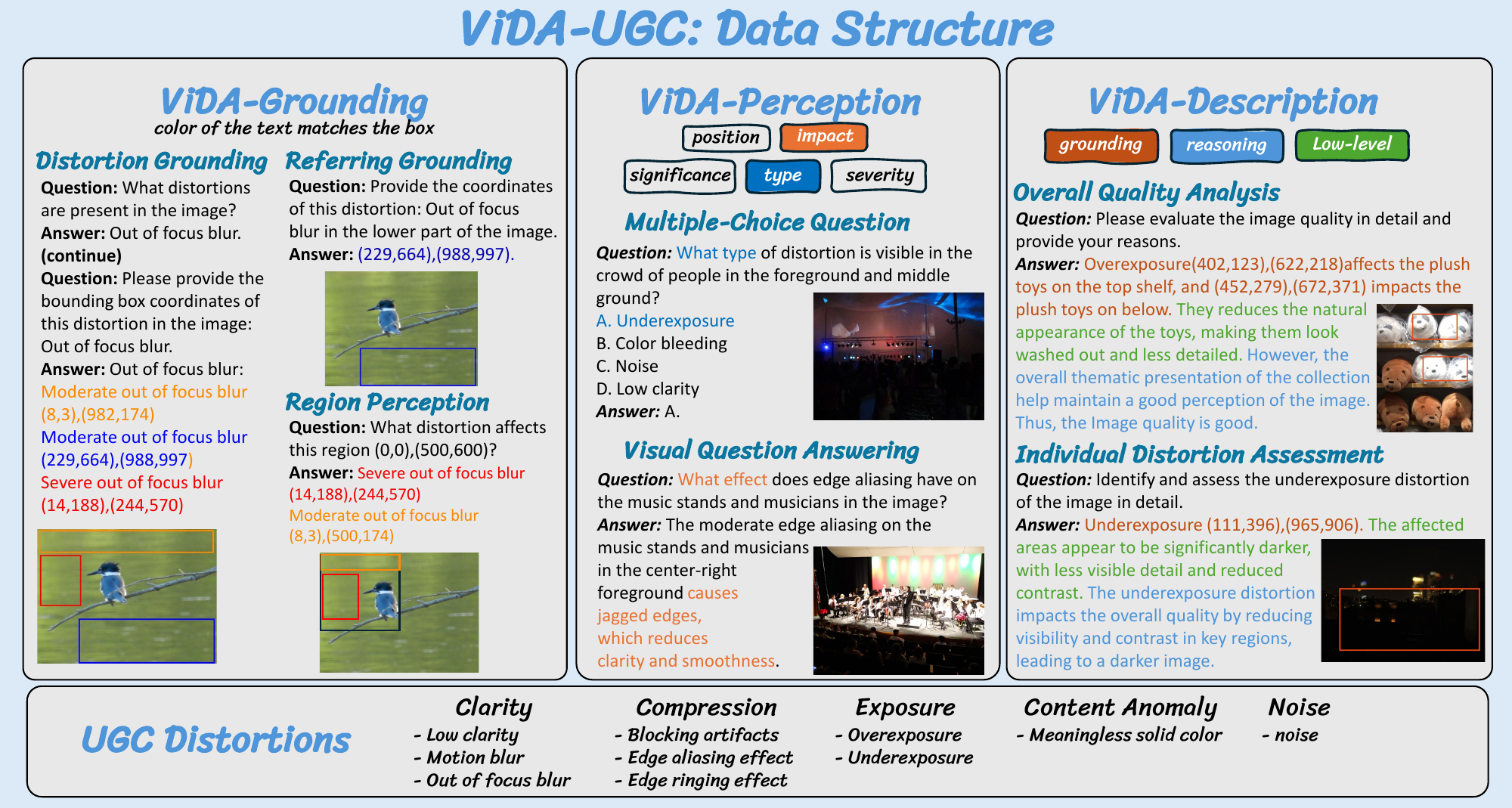}
    \caption{An illustration of the ViDA-UGC data structure. This dataset focuses on 10 common UGC distortions and is split into three sub-datasets: 1) ViDA-Grounding for fine-grained distortion grounding. This sub-dataset is divided into three tasks: distortion grounding, referring grounding, and region perception. 2) ViDA-Perception for detailed low-level perception. The questions in this sub-dataset have two formats and five concerns, focusing more on distortions. 3) ViDA-Description for reasoning quality description. Besides overall quality analysis, we add a new task, individual distortion assessment, which requires the model to assess a specific distortion in detail.}
    \label{fig:data}
    \vspace{-6pt}
\end{figure*}

\begin{abstract}
Recent advances in Multimodal Large Language Models (MLLMs) have introduced a paradigm shift for Image Quality Assessment (IQA) from unexplainable image quality scoring to explainable IQA, demonstrating practical applications like quality control and optimization guidance.
However, current explainable IQA methods not only inadequately use the same distortion criteria to evaluate both User-Generated Content (UGC) and AI-Generated Content (AIGC) images, but also lack detailed quality analysis for monitoring image quality and guiding image restoration.
%
In this study, we establish the first large-scale \textbf{Vi}sual \textbf{D}istortion \textbf{A}ssessment Instruction Tuning Dataset for \textbf{UGC} images, termed \textbf{ViDA-UGC}, which comprises 11K images with fine-grained quality grounding, detailed quality perception, and reasoning quality description data. This dataset is constructed through a distortion-oriented pipeline, which involves human subject annotation and a Chain-of-Thought (CoT) assessment framework. This framework guides GPT-4o to generate quality descriptions by identifying and analyzing UGC distortions, which helps capturing rich low-level visual features that inherently correlate with distortion patterns.
%
%
Moreover, we carefully select 476 images with corresponding 6,149 question answer pairs from ViDA-UGC and invite a professional team to ensure the accuracy and quality of GPT-generated information. The selected and revised data further contribute to the first UGC distortion assessment benchmark, termed \textbf{ViDA-UGC-Bench}.
%
Experimental results demonstrate the effectiveness of the ViDA-UGC and CoT framework for consistently enhancing various image quality analysis abilities across multiple base MLLMs on ViDA-UGC-Bench and Q-Bench, even surpassing GPT-4o. 
%
%
Project page: \href{https://whycantfindaname.github.io/ViDA-UGC}{https://whycantfindaname.github.io/ViDA-UGC}
\end{abstract}

\section{Introduction} \label{sec:intro}

Image Quality Assessment (IQA) seeks to evaluate image perceptual quality in alignment with the human visual system (HVS). With the rapid increase of digital content, IQA is becoming increasingly important in areas such as media streaming, user-generated photos, smartphone cameras, and the growing field of AI-generated content.
Attributed to the emergence of Multimodal Large Language Models (MLLMs) ~\cite{mplugowl,liu2023visual,qwenvl,internvl}, this field has witnessed a recent paradigm shift from unexplainable image quality scoring ~\cite{ssim, lpips, musiq, clipiqa} to explainable image quality assessment ~\cite{qbench, qinstruct, coinstruct, depictqav2}.
Explainable IQA leverages MLLMs' exceptional cross-modal capabilities to offer human-readable IQA with contextual descriptions for detailed visual quality analysis, which can guide quality control and restoration.
These works improve both the evaluation and explanatory capabilities of IQA systems, marking a new chapter for IQA.


Nevertheless, though existing MLLMs can basically reply to human queries regarding low-level visual aspects such as distortions (e.g., blur, noise, compression) and other features (e.g., color, lighting, composition) as introduced in Q-Bench \cite{qbench}, the accuracy of their responses remains unsatisfactory \cite{qbench, qbench+}. To achieve detailed quality analysis, many instruction tuning datasets have been established, aiming to enhance low-level perception \cite{huang2024visualcritic, qinstruct, coinstruct}, description \cite{huang2024visualcritic, qinstruct, coinstruct, depictqa, depictqav2}, and grounding capabilities \cite{chen2024grounding, liu2024grounding} for MLLMs. However, these datasets generally face two challenges: 
\begin{enumerate}
\item They inadequately adopt the same distortion criteria to evaluate both user-generated content (UGC) and AI-generated content (AIGC) images, while the focus of these two types of content differs significantly. UGC images encompass diverse capturing and processing conditions, often suffering from various distortions such as noise, blur, compression, \textit{etc}. In contrast, AIGC prioritizes aesthetics and text-image alignment, and some UGC distortions (e.g., out-of-focus blur) can even enhance the aesthetic appeal of AIGC images.
\item They are unable to simultaneously satisfy three tasks: \textbf{fine-grained distortion grounding}, \textbf{detailed low-level perception}, and \textbf{reasoning quality description}. 
Fine-grained distortion grounding involves precise localization of both global and local distorted regions in images.
Detailed low-level perception entails a multi-dimensional analysis of low-level visual attributes for both distortions and other features.
Reasoning quality description connects detailed low-level perception to quality reasoning via causal reasoning chains.
Previous works \cite{qinstruct,depictqav2,huang2024visualcritic} mainly focus on detailed low-level perception and reasoning quality description, which lack data for distortion grounding.
Recent works \cite{qground, liu2024grounding} segment distorted regions to achieve fine-grained grounding according to quality description data \cite{qinstruct}, but fail to leverage these regions to enhance MLLMs' perception and description capabilities. 
\end{enumerate}

The first challenge is difficult to resolve due to the divergent objectives between UGC-IQA and AIGC-IQA. Therefore, we narrow our focus to UGC-IQA on ten common UGC distortions, with the goal of achieving detailed quality analysis tailored to real-world UGC scenarios.

To solve the second challenge, we observe that humans have an interpretable pathway from distortion identification to quality reasoning: first localizing visual distortions, then perceiving low-level visual attributes, and finally providing a detailed logical feedback. Thus, we propose a \textbf{distortion-oriented dataset construction pipeline} to replicate this pathway, which consists of four steps.
\textbf{Step 1:} Sample images from UGC datasets and collect human-annotated distortion bounding boxes and mean opinion score (MOS).
\textbf{Step 2:} Generate textual descriptions of distortions and their visual attributes (position, severity, impact, significance) from GPT-4o based on human annotations.
\textbf{Step 3:} Propose a Chain-of-Thought (CoT) assessment framework to integrate human annotations and generated distortion textual descriptions.
\textbf{Step 4:} Convert these raw data to templated conversations for instruction tuning. 
Based on such a pipeline, we construct the first comprehensive~\textbf{Vi}sual~\textbf{D}istortion~\textbf{A}ssessment  Instruction
Tuning dataset for \textbf{UGC} images,~\textbf{ViDA-UGC}, which consists of 11,534 images, 36K distortion bounding boxes, and 534K instruction tuning data.
This dataset can be split into three sub-datasets: \textit{ViDA-Grounding} for fine-grained distortion grounding, \textit{ViDA-Perception} for detailed low-level perception, and \textit{ViDA-Description} for reasoning quality description, as illustrated in Figure \ref{fig:data}.
%
%

Although we integrate additional IQA expertise during GPT-based data generation to mitigate hallucinations, these generated data inevitably exhibit biases \cite{gptbias}. Meanwhile, the existing explainable IQA benchmark, Q-Bench \cite{qbench}, only evaluates MLLMs' general low-level visual perception and description capabilities, which is not comprehensive enough to fully evaluate the model's performance in the detailed quality analysis task. Thus, there is an urgent need for a more challenging and practical benchmark.
Therefore, we carefully select 476 images with 476 overall quality analyses from ViDA-Description, 2,567 multi-choice questions from ViDA-Perception, and 3,106 grounding data from ViDA-Grounding. Based on the collected images and data, a professional team consisting of image-processing researchers is responsible for ensuring the accuracy of GPT-4o generated information and revising the question-answer pairs, aiming to minimize GPT-4o biases. We term this benchmark as \textbf{ViDA-UGC-Bench}.
This benchmark tests the MLLMs from three explainable IQA tasks: distortion grounding, low-level perception, and quality description.
We observe significant performance drops in low-level perception and quality description tasks across MLLMs when compared to Q-Bench, as previous benchmarks comprise relatively simple questions and lack detailed and complex questions.
Extensive experimental results demonstrate the effectiveness of the ViDA-UGC for consistently enhancing various image quality analysis abilities across multiple base MLLMs on ViDA-UGC-Bench and Q-Bench, even surpassing GPT-4o.
Furthermore, our proposed CoT assessment framework can introduce causal reasoning chains and enrich low-level information in image quality description,  improving the description performance of existing MLLMs even without fine-tuning.

Our \textbf{contributions} can be summarized as follows:
\begin{enumerate}
\item We introduce a distortion-oriented dataset construction pipeline, which involves human subject annotation and a CoT framework. With this pipeline, we present ViDA-UGC, a comprehensive dataset, enabling detailed image quality analysis for UGC images.

\item We propose ViDA-UGC-Bench, which systematically evaluates MLLMs on fine-grained distortion grounding, detailed low-level perception, and reasoning quality description capabilities.

\item The proposed training-free CoT assessment framework is able to enhance the image quality description capabilities of current MLLMs and ensure high-quality data construction.
     
\end{enumerate}

\section{Related Work}
\label{sec:related}

\noindent \textbf{UGC-IQA Datasets and Benchmarks.}
Many IQA databases have been established to study the human perceptual quality characteristics of images. 
Early datasets typically require subjects to provide scores and derived MOS through aggregation \cite{livec,hosu2020koniq,lin2019kadid,spaq,lpips,pipal}. 
Correspondingly, the corresponding benchmarks generally evaluate the correlation between predicted scores from IQA models and these MOS values. Despite these datasets and benchmarks serving as important evaluation tools, the reliance on simple quality scores limits their interpretability.
Recently, Q-Bench \cite{qbench} first established a benchmark for evaluating MLLMs’ image general low-level perception and quality description abilities. Several works \cite{huang2024visualcritic, depictqa, qinstruct} further introduce large-scale low-level vision instruction tuning datasets to enhance both abilities. Recent advancements \cite{qground, liu2024grounding} pay more attention to local distortion identification and detailed quality analysis via image distortion grounding. 
However, these datasets adopt the same distortion criteria to evaluate both UGC and AIGC images, diminishing their value for UGC-IQA tasks.

\noindent \textbf{Instruction Tuning.}
Instruction tuning \cite{instruction-tuning} is a method proposed to improve the ability of MLLMs to follow instructions and enhance downstream task performance.
Previous models~\cite{qwenvl, liu2023visual, liu2024improved, mplugowl2, zhang2023internlm}, limited by their weak general low-level visual perception and understanding capabilities, required targeted instruction tuning data for improvement.
Recent MLLMs~\cite{internvl2.5, qwen2vl, internvl3}, leveraging superior architectures and larger training corpora, exhibit strong low-level visual abilities, achieving superior performance close to GPT-4o on Q-Bench.
However, existing MLLMs still fall short in detailed image quality analysis tasks. They generally lack fine-grained distortion grounding capabilities, while their perception and description capabilities remain at a superficial level, thereby calling for a comprehensive and detailed instruction tuning dataset to address these gaps.

\noindent \textbf{CoT Prompting.} 
CoT reasoning mechanisms \cite{wei2022chain,yao2023tree,besta2024graph} enable models to emulate human problem-solving by decomposing complex tasks into a sequence of manageable
sub-tasks, systematically constructing solutions. The intermediate reasoning steps or trajectories, referred to as the rationale, clarify the logical progression that underlies the model’s conclusions. Vanilla CoT \cite{wei2022chain} prompts models with \textit{``Think step by step.''}. Recent advancements \cite{guo2025deepseek, jaech2024openai} implicitly integrate this capability into cutting-edge systems through reinforcement learning. Our work applies CoT prompting to image quality description by deconstructing this task into step-wise reasoning chains, which can effectively analyze the impact of distortions and enrich low-level information.


\section{Proposed ViDA-UGC Dataset}
\label{sec:dataset}

In this section, we provide a comprehensive overview of our dataset construction, which serves as the foundation for detailed quality analysis.
First, we perform the in-lab subjective study, which is a preliminary step for subsequent processes and is introduced in the \textbf{\textcolor{red}{supplementary material}}.
Then, we discuss the step-wise reasoning stages of the proposed CoT assessment framework, which enhances the image quality description capabilities of current MLLMs and ensures high-quality data construction.
Finally, we dive into the distortion-oriented dataset construction pipeline. This pipeline integrates human subjects annotation and automated GPT-4o generation, thereby leveraging professional human IQA expertise and GPT's robust linguistic capabilities.
\begin{figure*}[t]
    \centering
    \includegraphics[width=1\textwidth]{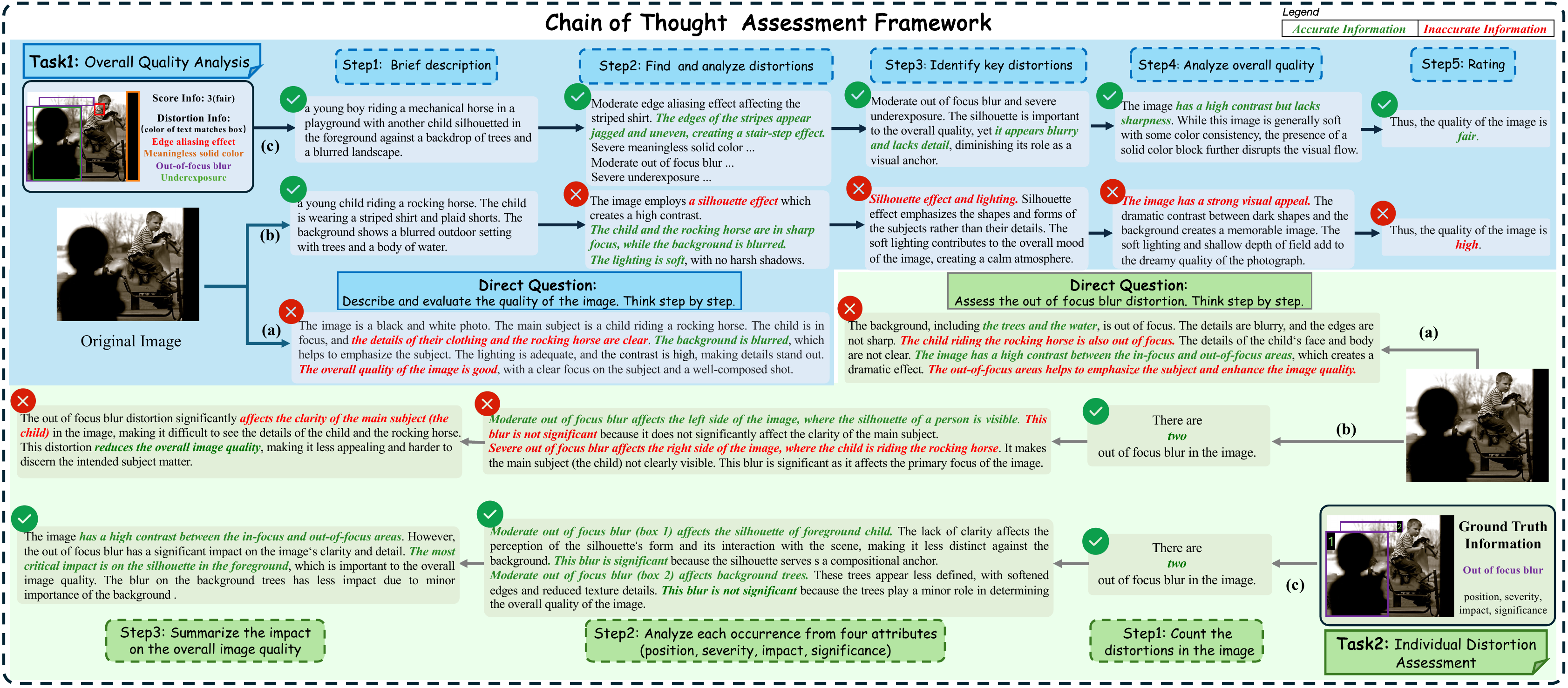}
    \caption{Overview of our CoT assessment framework. (a) Direct inference leads to inaccurate low-level information and superficial reasoning. (b) Standard CoT decomposes the task into reasoning steps but suffers from hallucination and inconsistent quality analysis. (c) Our approach integrates human expertise and ground-truth annotations to guide reliable step-wise reasoning and accurate quality prediction.}
    \label{fig:cot}
    \vspace{-4mm}
\end{figure*}

\subsection{CoT Assessment Framework}\label{sec:cot}
Considering the importance of UGC distortions in image quality description, we specifically decompose the quality description into two tasks: overall quality analysis and individual distortion assessment. We perform both tasks on example images from in-lab subjective study materials under different settings.
As illustrated in \Cref{fig:cot}(a), directly prompting existing MLLMs with \textit{``Think step by step."} leads to inaccurate low-level information and superficial reasoning.
To better explore the potential of existing models, we propose a CoT assessment framework that decomposes both tasks into step-wise reasoning chains. This approach enables detailed reasoning and supports granular analysis of distortions in intermediate steps, thereby enriching low-level information.
For overall quality analysis, the model is prompted to engage in a ``reasoning process" as Task 1 in \Cref{fig:cot}(b), which explicitly emulates the human quality analysis process from high-level to low-level:
1) Obtain a general impression of the image content;
2) Find and analyze the distortions to gather rich low-level information;
3) Further identify the key distortions; 
4) Analyze overall quality based on previous analysis; and
5) Rate the image quality.
The second to fourth steps include a comprehensive analysis of distortion patterns.
We also refine the second step to enable the model to perform individual distortion assessment, which requires a detailed analysis for a specific distortion as Task 2 in \Cref{fig:cot}(b).
Similar to overall quality analysis, this process explicitly guides the model from shallow to deep to provide a detailed analysis based on the defined distortion attributes.
This framework encourages the model to generate more detailed reasoning processes while better leveraging its pre-trained low-level knowledge, thereby yielding a notable performance improvement across both quality description tasks.

However, CoT framework remains constrained by the pre-trained low-level knowledge of MLLMs.
The multi-step process introduces error propagation issues \cite{yao2023react} and impacts reasoning efficiency.
If the model falls short in fine-grained distortion grounding and detailed low-level perception, failures in previous steps would result in incorrect analysis and ratings. As shown in \Cref{fig:cot}(a) and (b), the errors made by the model under these two prompt settings are highly similar, revealing inherent limitations in its capabilities.
This highlights the need for an end-to-end model with a strong distortion assessment capability to perform reasoning quality description for detailed quality analysis.

\subsection{Distortion-Oriented Dataset Construction Pipeline}\label{sec:pipeline}
\begin{figure}[t]
    \centering
    \includegraphics[width=0.46\textwidth]{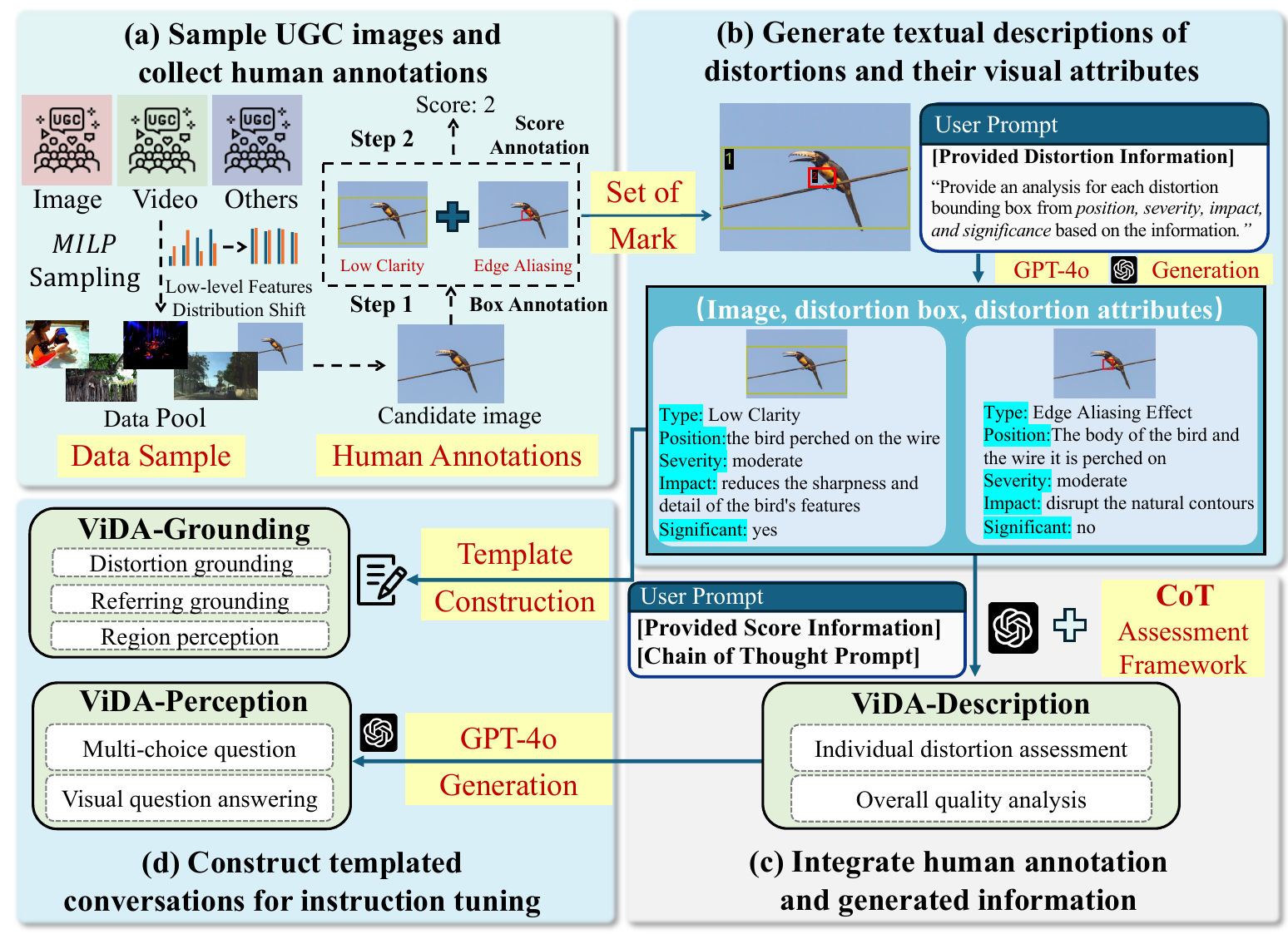}
    \caption{Overview of the ViDA-UGC construction pipeline. This pipeline includes four steps: In (a), MILP sampling strategy ensures approximately uniform distribution for sampled images across low-level features. Human subjects further annotate distortion boxes and image quality scores. In (b), we mark each box with a number and integrate IQA expertise to GPT prompt. GPT-4o then outputs textual descriptions of distortions and their visual attributes. In (c), the generated textual descriptions, together with human annotations and IQA expertise, serve as ground truth information in the proposed CoT framework, which is exploited by GPT-4o to generate quality description data. In (d), the quality description data are transformed into distortion-related VQA and MCQ via GPT-4o. Moreover, the generated distortion attributes in (b) are converted into grounding data using pre-defined chat templates. }
    \vspace{-4mm}
    \label{fig:anno}
\end{figure}
\begin{figure*}[t]
    \centering
    \includegraphics[width=01\textwidth]{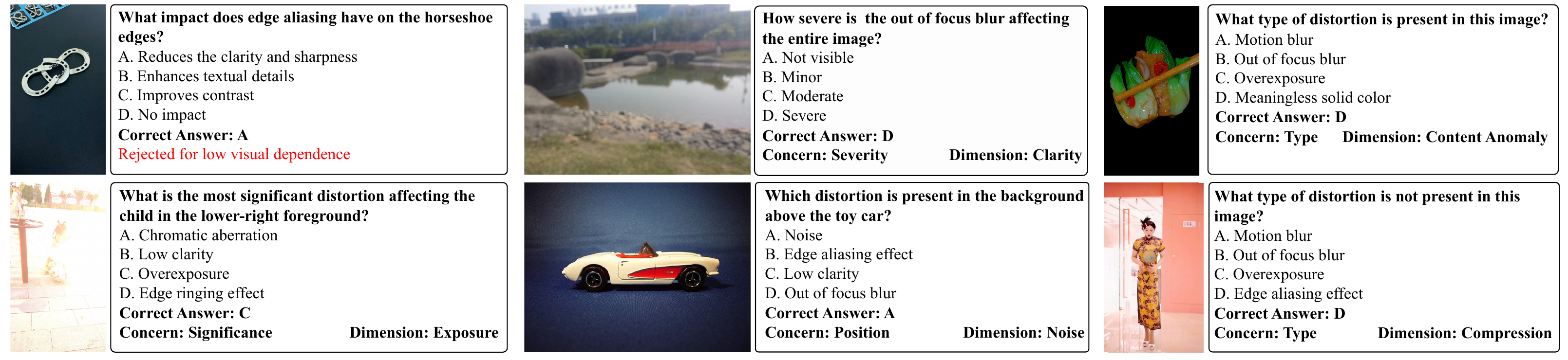}
    \caption{We show one rejected example, several accepted examples with different concerns and dimensions in ViDA-UGC-Bench.}
    \label{fig:bench}
    \vspace{-4mm}
\end{figure*}
The CoT assessment framework has demonstrated that existing MLLMs possess latent capabilities for detailed quality analysis but lack fine-grained distortion grounding and detailed low-level perception capabilities. In the following sections, we employ instruction tuning to explicitly bridge these capabilities. For this purpose, we construct a visual distortion assessment instruction tuning dataset for UGC images.
\Cref{fig:anno} presents the data construction pipeline, which is introduced step by step below.

\noindent \textbf{Step 1: Sample UGC Images and Collect Human Annotations.}
To ensure the dataset's diversity, we collect images from several UGC datasets and employ the improved data shaping method, Mixed Integer Linear Programming (MILP), as proposed in \cite{han2024evalmuse}, to sample part of images.
%
The final dataset consists of 11,534 images, each of which is further annotated by five human subjects.

For each image, quality scores and distortion bounding boxes are collected and cleaned.
Quality scores are invalidated and re-annotated if the difference between the maximum and minimum scores exceeds 1; otherwise, the MOS is calculated as the average of five scores.
Bounding boxes of each distortion are split into global and local types (Area Ratio threshold is set to 0.7). If global boxes exist, the largest global box is kept, and other boxes are discarded. Otherwise, Non-Maximum Suppression~\cite{nms} is applied to suppress excessive overlap between local boxes.
This process yields an average of 3.6 bounding boxes and one MOS score for each image.


\noindent \textbf{Step 2: Generate Low-Level Information from GPT-4o.}
Previous approaches \cite{qinstruct,huang2024visualcritic,depictqav2} have explored two methods for quality description: human annotations and GPT generation. However, human annotations provide limited low-level information, while GPT often makes inaccurate responses. We attribute these issues to  1) the former overly relies on the professionalism of human subjects, which is hindered by the scarcity of human experts and high annotation costs; and 2) the latter suffers from hallucinations, a common challenge in MLLMs. Current works adopt Retrieval-Augmented Generation (RAG) \cite{lewis2020retrieval}, which is a framework for integrating external knowledge retrieval for MLLMs, to enhance generative accuracy and solve hallucinations.
Following this idea, we integrate IQA expertise into GPT prompts. This helps GPT respond with more accurate low-level information as shown in \Cref{fig:cot}(c).
Specifically, we first define five attributes for each distortion box: type, position, severity, impact, and significance.
Based on the set-of-mark strategy ~\cite{setofmark}, we add visual marks of distortion regions on top of the image. This image, aligning with the corresponding textual distortion definition, is provided to GPT-4o for generating the triplet samples (image, distortion boxes, distortion attributes).

\noindent \textbf{Step 3: Integrate Human Annotation and Generated Information.}
To obtain correct and high-quality description data for each image, we adopt the proposed CoT assessment framework and integrate MOS, rating criteria, and corresponding distortion triplets into GPT prompt. 
By leveraging the rich low-level information derived from the provided texts, GPT-4o produces a logical reasoning process for both overall quality analysis and individual distortion assessment under the CoT framework, while ensuring the correctness of the analysis results.
We further integrate distortion grounding into the description with an interleaved format ``[distortion](bounding box)'' as ViDA-Description in \Cref{fig:data}, to avoid the mismatch between distortion grounding and quality description.

\noindent \textbf{Step 4: Convert Raw Data to Templated Conversations.}
While these reasoning quality descriptions form a key subset for enhancing MLLMs’ description ability, the full dataset is designed to unlock broader capabilities.
For low-level perception, we use GPT-4o to transform quality descriptions into distortion-related visual question answering (VQA) and multiple-choice questions (MCQ).
Unlike Q-Instruct, which focuses on question types like (what, how, yes-or-no), we target distortion attributes (type, position, severity, impact, significance) and other low-level features (e.g., lighting, composition).

For fine-grained distortion grounding, we divide this task into three sub-tasks:  distortion grounding, referring grounding, and region perception. Task data is structured in the form of pre-defined chat templates.
In the distortion grounding task, the model outputs the bounding boxes of required distortions, which are similar to the object detection task.
In the referring grounding task, the model receives target distortion information and outputs the corresponding box.
The target of this task is originally high-level objects, and we are the first to transfer the task from high-level object grounding to low-level distortion grounding.
In region perception, the model is offered with a bounding box of the region of interest (RoI) and locates distortions within the RoI \cite{chen2024seagull}.
To prioritize real-world applicability and generalization, these grounding sub-tasks are tailored for diverse queries from users, which are more complex and fine-grained than just asking for distortion locations.
We provide instances of these instruction tuning data in \Cref{fig:data}. Detailed statistics of ViDA-UGC and comparison between other IQA datasets are provided in the \textbf{\textcolor{red}{supplementary material}}.

\section{Proposed ViDA-UGC-Bench}

\label{sec:benchmark}
In this section, we introduce ViDA-UGC-Bench. We carefully select 476 images with their distortion triplet samples, 476 overall quality analysis from ViDA-Description, 2,567 multi-choice questions from ViDA-Perception, and 3,106 grounding data from ViDA-Grounding. Based on the collected images and data, a professional team consisting of image-processing researchers is responsible for revising the question-answer pairs and ensuring the accuracy of low-level information, aiming to minimize GPT-4o biases. While questions from existing works \cite{qbench, qbench+, depictqav2} are relatively straightforward and focus on factual elements, our revised questions are more challenging and practical, requiring detailed distortion understanding and covering concerns directly related to distortion attributes. Examples are shown in \Cref{fig:bench}. We show one rejected GPT-generated question, as its options are so simplistic that they can be selected without referring to the image. Other five accepted questions vary in difficulty. The hence-constructed ViDA-UGC-Bench covers all ten UGC distortions and tests MLLMs from three core IQA tasks: distortion grounding, low-level perception, and quality description. Detailed evaluation settings are provided in the \textbf{\textcolor{red}{supplementary material}}.

\section{Experiment}
\label{sec:experiment}
\subsection{Experimental Setups}

\noindent \textbf{Baselines.}
Our baseline models include Qwen-VL-Chat \cite{qwenvl}, Qwen2VL-7B-Instruct \cite{qwen2vl}, InternVL2.5-8B \cite{internvl2.5}, and InternVL3-8B \cite{internvl3}. We evaluate their distortion grounding, low-level perception, and quality description capabilities after instruction tuning using the Q-Instruct \cite{qinstruct} and our ViDA-UGC as training dataset, respectively. 

\noindent \textbf{Tasks for Evaluation.}
We use Q-Bench \cite{qbench} as the benchmark for quantitatively evaluating low-level perception ability. Since the \textbf{LLDescribe} dataset in Q-Bench is not publicly available, we select 100 images from Q-Pathway in descending order of the number of human-written quality descriptions and remove the quality description data from Q-Instruct to prevent test data leakage.
To evaluate detailed visual quality analysis abilities, we use the proposed ViDA-UGC-Bench.
\subsection{Main Results}
In this part, we conduct experiments to prove the effectiveness of our ViDA-UGC, CoT framework, and ViDA-UGC-Bench on three explainable quality tasks: low-level perception, quality description, and distortion grounding. Qualitative results and more detailed analysis about how our model surpasses GPT-4o are provided in the \textbf{\textcolor{red}{supplementary material}}.

\noindent \textbf{Perception.} 
As shown in \Cref{tab:perception}, baseline models show a 29\% average performance decline on ViDA-UGC-Bench compared to Q-Bench. This is because our ViDA-UGC-Bench introduces more detailed and challenging tasks than Q-Bench, which underscores limitations in existing MLLMs for detailed low-level perception. Training with ViDA-UGC consistently boosts MLLMs’ capabilities in both general and detailed low-level perception, whereas Q-Instruct fails to achieve comparable improvements. For Qwen2-VL-7B and InternVL3-8B, which achieve overall accuracy close to GPT-4o on Q-Bench, training with Q-Instruct degrades their performance, while their ViDA-UGC-tuned versions demonstrate improvements, with Qwen2-VL even surpassing GPT-4o's performance. This suggests that stronger baselines may already possess strong general perception capabilities that can be disrupted if the training dataset lacks valuable IQA expertise. For weaker baselines like Qwen-VL, ViDA-UGC still empowers them with significantly improved performance than Q-Instruct. Finetuning on Q-Instruct only slightly improves the overall performance of the weakest baseline Qwen-VL-Chat but degrades three others, while finetuning on ViDA-UGC significantly enhances baselines' ability to correctly perceive distortions from four attributes. Since question-answer pairs in Q-Instruct are generated by GPT-4o from quality descriptions in Q-Pathway, we infer that these human-written descriptions lack sufficient low-level details for GPT-4o, so these question-answer pairs are not effective for models whose performance is close to GPT-4o. 
In summary, ViDA-UGC enhances low-level perception abilities for MLLMs, while ViDA-UGC-Bench ensures robust evaluation across distortion types.

\begin{table*}[h!]
    \centering
    \renewcommand\arraystretch{1.16}
    \renewcommand\tabcolsep{4pt}
    \caption{Comparison of the \textbf{low-level Perception} ability among baseline MLLMs, Q-Instruct-\textit{tuned} versions, and \textbf{ViDA-UGC}-\textit{tuned} versions on both Q-Bench and ViDA-UGC-Bench. For each baseline model, the highest score is highlighted in \textbf{bold}. Here, we only present results for questions of distortion\textit{(Dist)} or in-context distortion\textit{(I-C Dist)} in Q-Bench. \textcolor{red}{Full results are available in the supplementary material.}}
    \resizebox{0.85\linewidth}{!}{\begin{tabular}{l|l|ccc|ccccc}
    \toprule
    \multirow{2}{*}{\textbf{Model} \textit{(variant)}} & \multirow{2}{*}{Training Dataset} & \multicolumn{3}{c}{\textbf{Q-Bench}}& \multicolumn{5}{c}{\textbf{ViDA-UGC-Bench}} \\ \cline{3-10} & & \textit{Dist}& \textit{I-C Dist}& \textit{Overall}& \textit{Type}& \textit{Position}&\textit{Severity}& \textit{Significance}& \textit{Overall}\\ \hline
    \multirow{4}{*}{Qwen-VL-Chat} 
        & \textit{no} (Baseline)& 50.78\% & 53.62\% & 56.39\% & 31.91\%& 37.83\%&31.72\%& 36.64\%& 34.84\%\\
        & Q-Instruct& 70.43\% & 74.34\% & 71.51\% & 28.50\%& 35.79\%&31.72\%& 47.12\%& 35.88\%\\
 & \textbf{ViDA-UGC}& \textbf{78.60\%}& \textbf{79.93\%}& \textbf{73.98\%}& \textbf{60.27\%}& \textbf{54.44\%}&\textbf{74.37\%}& \textbf{69.49\%}& \textbf{63.34\%}\\ \hdashline
    \multirow{4}{*}{Qwen2-VL-7B} 
        & \textit{no} (Baseline)& 75.49\% & 75.66\% & 77.19\%& 43.43\%& 43.53\%&51.26\%& 54.15\%& 47.53\%\\
        & Q-Instruct& 75.68\%& 77.63\%& 76.92\% & 34.12\%& 42.00\%&51.47\%& 52.08\%& 44.14\%\\
 & \textbf{ViDA-UGC}& \textbf{87.16\%}& \textbf{85.20\%}& \textbf{80.6\%}& \textbf{76.37\%}& \textbf{61.17\%}&\textbf{80.46\%}& \textbf{72.20\%}& \textbf{71.45\%}\\ \hdashline
    \multirow{4}{*}{InternVL2.5-8B} 
        & \textit{no} (Baseline)& 69.07\% & 70.07\% & 73.98\% & 37.08\%& 43.78\%&44.96\%& 49.04\%& 43.51\%\\
        & Q-Instruct& 76.65\% & 76.64\% & 76.45\% & 29.99\%& 38.32\%&68.07\%& 45.53\%& 43.40\%\\
 & \textbf{ViDA-UGC}& \textbf{87.74\%}& \textbf{84.54\%}& \textbf{79.33\%}& \textbf{81.24\%}& \textbf{62.06\%}&\textbf{82.14\%}& \textbf{78.27\%}& \textbf{74.80\%}\\ \hdashline
    \multirow{4}{*}{InternVL3-8B} 
        & \textit{no} (Baseline)& 70.43\% & 71.38\% & 74.72\% & 43.57\%& 47.08\%&47.06\%& 52.08\%& 47.37\%\\
        & Q-Instruct& 70.43\%& 75.99\%& 73.18\%& 29.10\%& 34.90\%&53.15\%& 47.28\%& 39.77\%\\
 & \textbf{ViDA-UGC}& \textbf{82.49\%}& \textbf{79.93\%}& \textbf{77.19/\%}& \textbf{82.87\%}& \textbf{61.80\%}&\textbf{78.15\%}& \textbf{72.52\%}& \textbf{73.00\%}\\ \hdashline
 GPT-4o& \textit{no} (Zero-shot)& 75.14\% & 78.10\%& 78.60\%& 46.38\%& 60.28\%& 53.57\%& 59.58\%&55.20\%\\
      \bottomrule
    \end{tabular}}
    \label{tab:perception}
\end{table*}

\begin{table*}[h!]
    \centering
    \caption{Comparison of the \textbf{Quality Description} ability among baseline MLLMs, Q-Instruct-\textit{tuned} versions, and \textbf{ViDA-UGC}-\textit{tuned} versions. For the training dataset with *, results are obtained under the prompt from Q-Instruct: \textit{``Describe and evaluate the quality of the image. Think step by step."}. For the datasets with †, results are obtained under our proposed CoT framework. For each baseline model, the highest score is highlighted in \textbf{bold} and the second highest score is marked in \textcolor{blue}{blue}.}
    \label{tab:description}
    \resizebox{0.9\linewidth}{!}{\begin{tabular}{l|l|cccc|cccc}
    \toprule
    \multirow{2}{*}{\textbf{Model} \textit{(variant)}} & \multirow{2}{*}{Training Dataset} & \multicolumn{4}{c}{\textbf{Q-Bench}}& \multicolumn{4}{c}{\textbf{ViDA-UGC-Bench}}\\ \cline{3-10}& & \textit{completeness} & \textit{precision}  &\textit{relevance}  & \textit{overall}& \textit{completeness} & \textit{precision} &reasoning  & \textit{overall}\\ \hline
    \multirow{4}{*}{Qwen-VL-Chat} 
        & \textit{no} (Baseline)* & 0.72 & 0.53  &1.86  & 3.11& 0.2& 0.58&1.56 & 2.34\\
        & \textit{no} (Baseline)† & 0.78& 0.55 &\textcolor{blue}{1.99}& 3.32& 0.5& 0.77&\textcolor{blue}{2.13} & 3.4\\
        & Q-Instruct* & \textcolor{blue}{0.97}&\textcolor{blue}{0.76} &1.95& \textcolor{blue}{3.68}& \textcolor{blue}{0.58}& \textcolor{blue}{0.93}&1.98 & \textcolor{blue}{3.49}\\
 & \textbf{ViDA-UGC}† & \textbf{1.07}& \textbf{0.74} &\textbf{1.99}& \textbf{3.80}& \textbf{1.33}& \textbf{1.14}&\textbf{2.89} & \textbf{5.36}\\ \hdashline
    \multirow{4}{*}{Qwen2-VL-7B} 
        & \textit{no} (Baseline)* & 1.30& 1.06 &\textbf{2}& 4.36& 0.70& 0.73&2.78 & 4.21\\
        & \textit{no} (Baseline)† & \textcolor{blue}{1.36}& 1.28 &\textbf{2}& \textcolor{blue}{4.64}& \textcolor{blue}{0.74}& 0.92&\textcolor{blue}{2.88} & \textcolor{blue}{4.54}\\
        & Q-Instruct* & 1.15& \textbf{1.35} &\textbf{2}& 4.50& 0.69& \textcolor{blue}{1.14}&1.78 & 3.61\\
 & \textbf{ViDA-UGC}† & \textbf{1.40}& \textcolor{blue}{1.30} &\textbf{2}& \textbf{4.70}& \textbf{1.49}& \textbf{1.34}&\textbf{2.95} & \textbf{5.78}
\\ \hdashline
    \multirow{4}{*}{InternVL2.5-8B} 
        & \textit{no} (Baseline)* & 0.96& 0.72 &1.83& 3.51& 0.74& 0.74&2.76 & 4.24\\
        & \textit{no} (Baseline)† & \textcolor{blue}{1.15}& 1.03 &\textcolor{blue}{1.94}& \textcolor{blue}{4.12}& \textcolor{blue}{0.75}& 1.25&\textcolor{blue}{2.90} & \textcolor{blue}{4.9}\\
        & Q-Instruct* & 0.97& \textcolor{blue}{1.22} &1.93& \textcolor{blue}{4.12}& 0.63& \textcolor{blue}{1.28}&1.63 & 3.54\\
 & \textbf{ViDA-UGC}† & \textbf{1.22}& \textbf{1.23} &\textbf{1.99}& \textbf{4.44}& \textbf{1.46}& \textbf{1.32}&\textbf{2.94} & \textbf{5.72}\\ \hdashline
    \multirow{4}{*}{InternVL3-8B} 
        & \textit{no} (Baseline)* & 1.10& 0.93 &1.86& 3.89& 0.93& 0.96&2.95 & 4.84\\
        & \textit{no} (Baseline)† & \textcolor{blue}{1.27}& \textcolor{blue}{1.34} &\textbf{2}& \textcolor{blue}{4.61}& \textcolor{blue}{0.96}& \textcolor{blue}{1.28}&\textcolor{blue}{2.98} & \textcolor{blue}{5.22}\\
        & Q-Instruct* & 0.98& 1.32 &1.95& 4.25& 0.64& 1.25&1.63 & 3.52\\
 & \textbf{ViDA-UGC}† & \textbf{1.34}& \textbf{1.35} &\textcolor{blue}{1.99}& \textbf{4.68}& \textbf{1.51}& \textbf{1.36}&\textbf{3} & \textbf{5.87}\\
    \bottomrule
\end{tabular}}
\end{table*}
\begin{table}\small
    \centering
    \renewcommand\tabcolsep{3pt}  
    \caption{Comparison of the \textbf{Referring Grounding} ability between baselines and \textbf{ViDA-UGC}-\textit{tuned} versions. The highest score is highlighted in \textbf{bold} and the highest improvement is marked in \textcolor{red}{red}. Response rate reflects the frequency of successfully returning bounding box coordinates.}
    \label{tab:referring_grounding}  
    \resizebox{0.9\linewidth}{!}{
    \begin{tabular}{l|c c c}  
        \toprule  
        Method & Response Rate & Acc$_{0.5}$ & mIoU \\
        \hline  
        Qwen-VL-Chat       & 1    & 32.4 & 37.3 \\
        Qwen-VL-Chat-ViDA  & 1    & 41.3$_{(+8.9)}$  & 43.4$_{(+6.1)}$  \\
        \hdashline  
        Qwen2-VL-7B        & 0.79 & 24.9 & 29.8 \\
        Qwen2-VL-7B-ViDA   & 0.99 & 42.1$_{(+17.2)}$ & 45.2$_{\textcolor{red}{(+15.4)}}$ \\
        \hdashline
        InternVL2.5-8B     & 0.98 & 29.0 & 37.0 \\
        InternVL2.5-8B-ViDA& 1    & 43.3$_{(+14.3)}$ & 46.4$_{(+9.4)}$ \\
        \hdashline
        InternVL3-8B       & 0.95 & 25.8 & 33.5 \\
        InternVL3-8B-ViDA  & 1    & \textbf{44.2}$_{\textcolor{red}{(+18.4)}}$ & \textbf{47.0}$_{(+13.5)}$ \\
        \bottomrule  
    \end{tabular}}
\end{table}
\noindent \textbf{Description.} 
As shown in \Cref{tab:description}, the proposed distortion-oriented CoT empowers MLLMs with stronger quality description ability, achieving better results than Q-Insturct-\textit{tuned} version for Qwen2-VL, InternVL2.5, and InternVL3. This framework consistently improves all description metrics across both benchmarks, with notable gains in relevance and reasoning. These results demonstrate that existing MLLMs possess latent capabilities for detailed quality analysis, which can be effectively unlocked through an appropriate methodology. Specifically, finetuning on Q-Instruct yields marginal improvements in completeness and precision relative to ground-truth information, yet it diminishes reasoning scores. This observation highlights a key limitation of Q-Instruct: their quality descriptions lack a robust reasoning process. In contrast, finetuning on ViDA-UGC strengthens description performance, not only activating MLLMs’ inherent capabilities but also enhancing their low-level vision knowledge for analysis.


\noindent \textbf{Grounding.} 
We compare the distortion referring grounding between baseline MLLMs and ViDA-UGC-\textit{tuned} versions in \Cref{tab:referring_grounding}. These four baselines perform well on high-level benchmarks \cite{refcoco+,refcoco}, achieving nearly 90 Acc$_{0.5}$. When transferring to low-level distortion grounding, they drop significantly. Moreover, Qwen2-VL-7B fails to respond correctly sometimes, reflected by only a 0.72 response rate. Finetuning on ViDA-UGC not only makes them consistently output correct results but also improves grounding precision.

\section{Conclusion}
\label{sec:conclusion}
In this work, we dive into detailed visual quality analysis with MLLMs for explainable IQA. We introduce a comprehensive distortion assessment dataset based on a distortion-oriented construction pipeline. During construction, we propose a training-free distortion-oriented CoT framework that effectively enhances the reasoning process in MLLM's quality descriptions. Additionally, we design ViDA-UGC-Bench, a benchmark focusing on distortion assessment, revealing significant limitations of existing MLLMs in detailed quality analysis. Experimental results showcase that ViDA-UGC allows MLLMs to unlock three key tasks toward detailed quality analysis.

%
\bibliography{aaai2026}

\clearpage


\section*{Supplementary material}

# ViDA-UGC: Detailed Image Quality Analysis via Visual Distortion Assessment for UGC Images

*——————— Supplementary Material ———————*

## Overview

This document provides supplementary material for the main paper. The supplementary material contains four main parts:

- More details of ViDA-UGC Dataset, including data sources, in-lab subjective study, and dataset statistics.
- More details of ViDA-UGC-Bench, including evaluation settings on perception, description, and grounding tasks.
- More details of experiments, including model and training details, detailed experimental results, and analysis.
- Qualitative results, including low-level perception and quality description.

## More Details of ViDA-UGC Dataset

Table 1: The image sources.

| Image Sources | Original Images(Videos) | Selected Samples |
|---|---|---|
| KonIQ-10K (Hosu et al. 2020) | 10,373 | 1125 |
| SPAQ (Fang et al. 2020) | 11,125 | 1146 |
| LIVE-FB (Ghadiyaram and Bovik 2015) | 39,810 | 2039 |
| LIVE-itw (Ghadiyaram and Bovik 2015) | 1,169 | 77 |
| DIV2K (Agustsson and Timofte 2017) | 2000 | 328 |
| DPED (Ignatov et al. 2017) | 30000 | 674 |
| CVD2014 (Nuutinen et al. 2016a) | 234 | 12 |
| CVQAD (Antsiferova et al. 2022) | 1023 | 70 |
| ICME2021-UGC (Wang et al. 2021) | 7200 | 1217 |
| Dover-UGC (Wu et al. 2023) | 3590 | 368 |
| TaoLive (Zhang et al. 2023) | 3762 | 82 |
| Others | - | 4396 |

### Data Sources

We choose diverse raw datasets, including four in-the-wild IQA datasets (Fang et al. 2020; Hosu et al. 2020; Niu 2022; Ghadiyaram and Bovik 2015), two image super-resolution datasets (Ignatov et al. 2017; Agustsson and Timofte 2017), several video datasets (Nuutinen et al. 2016b; Wu et al. 2023; Zhang et al. 2023; Antsiferova et al. 2022), and other in-house data in Table 1. For video data, we extract the first frame from each video. To achieve a balance between high-quality and low-quality images, we set several feature dimensions including colorfulness, sharpness, light contrast, entropy, and scape mode. We calculate feature values and employ the improved data shaping method, Mixed Integer Linear Programming (MILP), as proposed in (Han et al.

| Score | Level | Definition | Example Images |
|---|---|---|---|
| 1 | Bad | The overall clarity of the picture is low, or its meaning is unclear, making it difficult to distinguish the content of the image. | |
| 2 | Poor | The clarity of the main subject in the frame is relatively low, but the edge contours can be distinguished. The entire image has low contrast or appears darkish, with obvious noise, coding compression issues, and so on. | |
| 3 | Fair | The main content of the frame is relatively clear, while the edge and background areas have distortions such as visual blur or dimness; examples include significant noise, visual blur, local light spots, edge sharpening, or significant blurriness of background textures. | |
| 4 | Good | The main subject of the image is clear, and the entire image has no obvious noise, visual blur, camera shake, or imaging light spots. However, the overall texture of the image has limited complexity and lacks rich details and layers; or there are slightly visible local artifacts or exposure issues. | |
| 5 | Excellent | The overall clarity and contrast of the frame are excellent, with high resolution, rich and clear textures and colors. There may be slight distortion issues, but they do not affect the overall quality | |

Figure 1: The rating criteria and example images for the overall quality score and corresponding quality level, which ranges from 1 to 5.

2024), to sample from original images. Finally, we collected 11,534 images.

### In-Lab Subjective Study

The subjective study is carried out in a well-controlled laboratory environment. A group of UGC-IQA human experts first establishes UGC distortion criteria, which contain five dimensions: *clarity, compression, exposure, content anomaly, noise*. Ten common UGC distortions across the five dimensions are chosen for detailed quality analysis. Training materials are developed based on professional IQA resources (ITU-R 2000; ISO 2005, 2015; ITU-R 2019) and domain expertise, consisting of two components: 1) the rating criteria for the overall quality score presented in Figure 5, and 2) the definition for each distortion type presented in Figure 1. Each component also includes numerous example images for reference. Before the subjective experiments, we provide these training materials to human subjects. Subjects are trained to annotate distortion bounding boxes on an image and assign quality scores to images within the range {1, 2, ..., 5} based on their annotations. To ensure reliabil-

Table 2: **Comparison of public IQA datasets and the proposed ViDA-UGC.** DG, RG, RP, mcq, and vqa respectively represent distortion grounding, referring grounding, region perception, multiple-choice question, and visual question answering.

| Dataset | Quality Scoring | Quality Grounding | | | low-level perception | | Quality Description | | |
|---|---|---|---|---|---|---|---|---|---|
| | MOS | DG | RG | RP | mcq | vqa | low-level related | with grounding | rich reasoning |
| Traditional IQA datasets | ✓ | ✗ | ✗ | ✗ | ✗ | ✗ | ✗ | ✗ | ✗ |
| Q-Instruct-200K (Wu et al. 2024b) | ✓ | ✗ | ✗ | ✗ | ✓ | ✓ | ✓ | ✗ | ✗ |
| DepictQA-Wild (You et al. 2024) | ✓ | ✗ | ✗ | ✗ | ✗ | ✓ | ✓ | ✗ | ✓ |
| QGround-100K (Chen et al. 2024a) | ✓ | ✓ | ✓ | ✗ | ✓ | ✓ | ✓ | ✗ | ✗ |
| **ViDA-UGC** | ✓ | ✓ | ✓ | ✓ | ✓ | ✓ | ✓ | ✓ | ✓ |

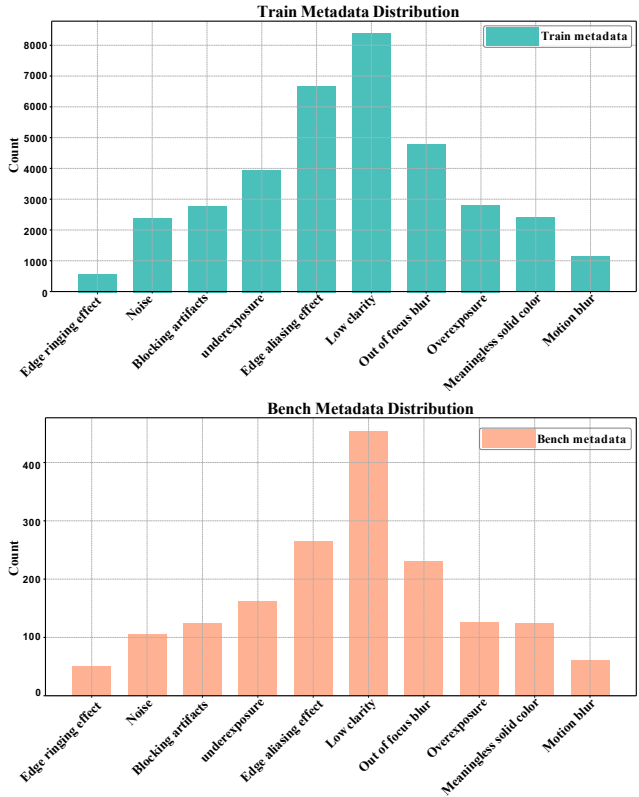

Figure 2: Distortion distributions. **Top:** ViDA-UGC. **Bottom:** ViDA-UGC-Bench.

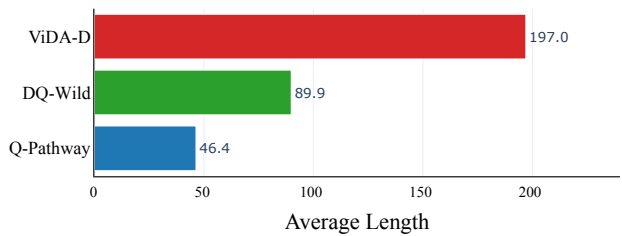

Figure 3: Comparison of length distributions of quality descriptions in ViDA-UGC, DepictQA-Wild, and Q-Instruct.

ity and consistency of the annotation, a qualification exam is conducted after the training session. A total of 300 images are selected as exam materials, including a subset of images with identical content but different distortion types to assess the annotators' discrimination ability. In addition, 30 images are duplicated within the exam set as a consistency check mechanism. An annotator is deemed qualified only if their annotations for the duplicated images are internally consistent and match the ground truth (GT) labels.

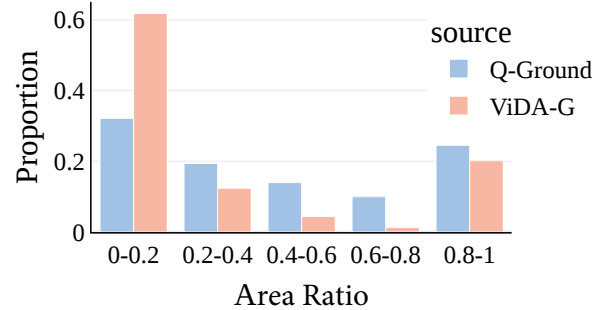

Figure 4: Comparison of area ratio distributions of boxes in ViDA-UGC and masks in Q-Ground.

## Data Statistics

**Distortion Type Distribution.** We select five distortion dimensions *(clarity, compression, exposure, content anomaly, noise)* as a basis to explore distortion assessment, which includes ten prevalent types of distortions. After collecting all the source data, we split the whole dataset into train and benchmark. As shown in Figure 2, the distribution trends of the distortion types in both subsets are consistently related, indicating that the benchmark set effectively reflects the feature distribution of the training set. The most common degradation type is "Low clarity", which occupies the highest proportion in both datasets. Followed by "Underexposure" and "Edge aliasing effect". Two of the rarest types are "Motion blur" and "Edge ringing effect".

**Quality Description Length Distribution.** The comprehensiveness of quality description data in ViDA-UGC is reflected in the average length: Figure 3 shows the distribution of average quality description lengths in ViDA-UGC, DepictQA-Wild (You et al. 2024), and Q-Instruct (Wu et al. 2024b), with ViDA-UGC containing a large number of detailed descriptions and achieving significantly longer average lengths.

**Bounding Box Area Ratio Distribution.** We compare the area ratios of distortion masks in Q-Ground (Chen et al. 2024a) and distortion boxes in ViDA-Grounding in Figure 4. Although Q-Ground uses a "smaller region first" principle to merge overlapping masks, our distortion boxes with an area ratio below 0.2 far outnumber Q-Ground's masks, highlight-

ing our emphasis on local and precise distortion regions.

Table 3: Completeness Evaluation Prompt. [MLLM_DESC] and [DISTORTION_INFO] are replaced by the output description from MLLMs and distortion attributes (e.g., type, position, impact, significance) in the ViDA-UGC-Bench.

> **Completeness Evaluation Prompt**
>
> User: You are an expert in image quality assessment. Your task is to evaluate the completeness of the MLLM description based on the detailed low-level visual information derived from the reference. Evaluate whether the description [MLLM_DESC] completely includes the low-level visual information in the reference distortion information [DISTORTION_INFO].
> Please rate score 3 for completely or almost completely including reference information, 0 for not including at all, 2 for including a large part of the information or a similar description, 1 for including a small part of the information or a similar description.
> Please only provide the result in the following format:
> Score:

## Dataset Adavantages

**Diversity.** Our dataset includes a wide range of common distortion types, covering various image quality issues encountered in real-world scenarios. Such diversity enables the model to learn a broader range of distortion features, enhancing its generalization ability.

**Real-World Representativeness.** While there is some imbalance in the distribution of distortion types in the dataset, we believe such imbalance mirrors the real-world distribution of image distortions. For example, from experience, low clarity is the most common distortion seen in real world, so it naturally should occupy a larger proportion of the dataset. Such dataset distribution helps the MLLMs learn more effectively and helps us evaluate MLLMs more closely in line with the real world.

**Task Complexity.** We compare ViDA-UGC with existing public IQA datasets in Table 2. Traditional IQA datasets rely on simple quality scores and lack interpretability. Earlier explainable IQA datasets (Wu et al. 2024b; You et al. 2024) introduce textual quality descriptions and perception questions, greatly improving interpretability. However, they lack quality grounding tasks, leading to a poor understanding of local distortions and less detailed distortion-related questions. Q-Ground introduces quality grounding on top of Q-Instruct but neglects perception and description enhancements. ViDA-UGC create a comprehensive dataset for three explainable IQA tasks by expanding grounding tasks, refining perception questions, and providing reasoning quality descriptions.

## More Details of ViDA-UGC-Bench

### Evaluation Settings

**Benchmark on Distortion Grounding Ability.** We measure the ViDA-Grounding performance for three grounding subtasks. (1) For distortion grounding, the model is required to output all distortion bounding boxes. We use COCO mAP (Lin et al. 2014) to test its performance. (2) For referring grounding, the model is given distortion type and location, arranged in '[distortion type] affecting [location]'. It is required to directly output corresponding box coordinates, and we use the conventional metric $Acc_{0.5}$ in referring expression comprehension (Kazemzadeh et al. 2014). (3) For region perception, the model receives a random box in the image and grounds distortions within the box. We also use COCO mAP as the evaluation metric.

**Benchmark on Low-Level Perception Ability.** We adopt a multiple-choice assessment framework following Q-Bench's methodology (Wu et al. 2024a) to evaluate four distortion attributes: type, severity, position, and significance. The impact attribute is deliberately excluded from our evaluation protocol due to its inherent descriptive subjectivity. We employ Accuracy as a metric. Denote the image tokens as `<image>`, the question as `<QUESTION>`, choices as `<CHOICE_i>`, the user prompt template for Multi-Choice Questions (MCQ) is listed in Table 4.

Table 4: MCQ Prompt. "{*Question*}" and "{*CHOICE*}" is replaced by user questions and options during training and inference.

> **MCQ Prompt**
>
> *USER: You are an expert in image quality assessment.*
> `<image>` `<QUESTION>`
> *Answer with the option's letter from the given choices directly.*
> *A.* `<CHOICE_A>`
> *B.* `<CHOICE_B>`
> *C.* `<CHOICE_C>`
> *D.* `<CHOICE_D>`

**Benchmark on Quality Description Ability.** We evaluate ViDA-Description performance for overall quality analysis. Following Q-Bench, we conduct a five-round GPT evaluation between ground truth distortion information in an image and the model-generated analysis under three dimensions. We pose the same prompt as defined in the following templates five times, taking the mean result of GPT's answers to determine the final outcome. The three dimensions are: (1) Completeness. More matched information with the golden information is encouraged. (2) Precision. The controversial information with the golden information is punished. (3) Reasoning. MLLMs' outputs should have coherent reasoning chains rather than simply stating low-level visual observations. Each dimension is scored among $[0, 1, 2, 3]$ and the weighted average is collected as the final score.

Table 5: Baseline MLLMs for instruction tuning.

| Month/Year | Model Name | Visual Backbone | Language Model | Image Preprocess |
|---|---|---|---|---|
| Jul/23 | Qwen-VL-Chat (Bai et al. 2023) | CLIP-ViT-bigG$^{\uparrow 448}$ | Qwen-7B | Resize to (448, 448) |
| Sep/24 | Qwen2-VL-7B-Instruct (Wang et al. 2024) | Custom ViT | Qwen2-7B | Naive Dynamic Resolution |
| Oct/24 | InternVL3-8B (Zhu et al. 2025) | InternViT-300M$^{\uparrow 448}$ | Qwen2.5-7B | Dynamic High-Resolution |
| Nov/24 | InternVL2.5-8B (Chen et al. 2024b) | InternViT-300M$^{\uparrow 448}$ | Internlm2.5-7B | Dynamic High-Resolution |

Table 6: Precision evaluation prompt.

> **Preciseness Evaluation Prompt**
>
> User: You are an expert in image quality assessment. Your task is to evaluate the preciseness of the MLLM description [MLLM_DESC] based on the detailed low-level visual information derived from the reference distortion information [DISTORTION INFO].
> This metric punishes low-level descriptions that significantly deviate from the reference distortion information(e.g., blur for clear, significant for insignificant, colorful for monotonous, noisy for clean, bright for dark), while minor discrepancies are not considered violations.
> Please rate score 3 for totally no controversial low-level description, 2 for a small number of controversial low-level descriptions compared to distortion information, 1 for a moderate number of controversial low-level descriptions, and 0 for a large number of controversial low-level descriptions.
> Please only provide the result in the following format: Score:

Table 7: Reasoning evaluation prompt.

> **Reasoning Evaluation Prompt**
>
> User: You are an expert in image quality assessment. Please Evaluate whether the reasoning in [MLLM DESC] demonstrates comprehensive technical analysis and logical coherence.
> Please rate score 3 for multi-stage reasoning with precise technical terms, score 2 for clear analysis with minor logic gaps, score 1 for basic observations with weak reasoning, and score 0 for irrelevant/illogical statements.
> Please only provide the result in the following format: Score:

We introduce user prompt templates for GPT-4o evaluation in Table 3, Table 6, and Table 7.

## More Details of Experiments

### Model and Training Details

We choose four baseline models with diverse meta structures and image preprocessing methods in Table 5. We finetune them with Q-Instruct and ViDA-587K using the same

Table 8: Comparison of the **Distortion Grounding and Region Perception** abilities between finetuned object detection methods and **ViDA-UGC**-*tuned* MLLMs. DES and DG are two methods we use for distortion grounding tasks. DES means we extract each distortion box from the model's overall quality description. DG means we directly perform the distortion grounding task. RP means region perception.

| Method | DES mAP | DG mAP | RP mAP |
|---|---|---|---|
| TOOD (Feng et al. 2021) | - | 29.2 | 36.2 |
| Co-DETR (Zong, Song, and Liu 2023) | - | 31.4 | 35.8 |
| Grounding Dino (Liu et al. 2024) | - | 37.1 | 42.2 |
| InternVL3-8B-ViDA | 17.4 | 19.7 | 30.4 |
| InternVL2.5-8B-ViDA | 15.9 | 20.0 | 32.3 |
| Qwen2-VL-7B-ViDA | 17.9 | 17.7 | 29.4 |
| Qwen-VL-Chat-ViDA | 15.5 | 16.8 | 27.1 |

settings, respectively and provide hyper-parameter configuration for baselines: Table 9 for Qwen-VL-Chat, Table 10 for Qwen2-VL-Instruct, Table 11 for InternVL2.5-8B, Table 12 for InternVL3-8B.

### Detailed Experimental Results And Analysis

**Perception Results.** The experimental results demonstrate a striking contrast in model performance between the two benchmarks, highlighting the unique challenges posed by each evaluation framework. Q-Bench primarily evaluates general perception abilities through diverse question types (Yes-or-No, What, How), while ViDA-UGC-Bench focuses specifically on distortion-oriented dimensions (Clarity, Compression, Exposure, Content) and assessment concerns (Noise, Type, Position, Severity, Significance) that are particularly relevant for user-generated content.

Performance on Q-Bench is in Table 13. We observe several noteworthy patterns:
1) For most baseline models, fine-tuning with ViDA-UGC consistently improves performance on overall accuracy, with particularly significant gains in distortion-related categories. This suggests that ViDA-UGC's distortion-oriented training enhances models' ability to identify and characterize image quality issues. However, ViDA-UGC-tuned MLLMs encounter severe performance degradation on Other and In-Context Other questions, which may concern low-level attributes as color, lighting, and composition.
2) Q-Instruct tuning shows inconsistent effects across different model architectures. For Qwen-VL-Chat and

Table 9: Hyper-parameter configurations for Qwen-VL-Chat.

| Hyper-parameter config | Value |
| --- | --- |
| Input image size | $448 \times 448$ |
| Image encoder | frozen |
| Adapter | frozen |
| LLM | Lora |
| Lora rank | 64 |
| Lora alpha | 128 |
| Optimizer | AdamW |
| Learning rate | 1e-4 |
| Weight decay | 0.05 |
| $(\beta_1, \beta_2)$ | $(0.9, 0.95)$ |
| Scheduler | WarmupCosineLR |
| Warm up ratio | 0.03 |
| ZeRO stage (deepspeed) | 3 |
| Precision | bfloat16 |
| Batch size (with accumulation) | $8 \times 16 \times 1$ |
| Total epochs | 3 |

Table 10: Hyper-parameter configurations for Qwen2-VL-7B-Instruct.

| Hyper-parameter config | Value |
| --- | --- |
| Input image size | Original |
| Image encoder | Lora |
| Adapter | Full |
| LLM | Lora |
| Lora rank | 64 |
| Lora alpha | 128 |
| Optimizer | AdamW |
| Learning rate | 1e-4 |
| Merger LR | 1e-5 |
| Vision LR | 1e-6 |
| Weight decay | 0.05 |
| $(\beta_1, \beta_2)$ | $(0.9, 0.95)$ |
| Scheduler | WarmupCosineLR |
| Warm up ratio | 0.03 |
| ZeRO stage (deepspeed) | 3 |
| Precision | bfloat16 |
| Batch size (with accumulation) | $16 \times 8 \times 1$ |
| Total epochs | 3 |

Table 11: Hyper-parameter configurations for InternVL2.5-8B.

| Hyper-parameter config | Value |
| --- | --- |
| Input image size | Original |
| Image encoder | Frozen |
| Adapter | Full |
| LLM | Lora |
| Lora rank | 64 |
| Lora alpha | 128 |
| Optimizer | AdamW |
| Learning rate | 1e-4 |
| Weight decay | 0.05 |
| $(\beta_1, \beta_2)$ | $(0.9, 0.95)$ |
| Scheduler | WarmupCosineLR |
| Warm up ratio | 0.03 |
| ZeRO stage (deepspeed) | 3 |
| Precision | bfloat16 |
| Batch size (with accumulation) | $8 \times 16 \times 1$ |
| Total epochs | 3 |

Table 12: Hyper-parameter configurations for InternVL3-8B.

| Hyper-parameter config | Value |
| --- | --- |
| Input image size | Original |
| Image encoder | Frozen |
| Adapter | Full |
| LLM | Lora |
| Lora rank | 64 |
| Lora alpha | 128 |
| Optimizer | AdamW |
| Learning rate | 1e-4 |
| Weight decay | 0.05 |
| $(\beta_1, \beta_2)$ | $(0.9, 0.95)$ |
| Scheduler | WarmupCosineLR |
| Warm up ratio | 0.03 |
| ZeRO stage (deepspeed) | 3 |
| Precision | bfloat16 |
| Batch size (with accumulation) | $8 \times 16 \times 1$ |
| Total epochs | 3 |

InternVL2.5-8B, Q-Instruct tuning yields modest improvements over baseline performance. However, for more advanced models like Qwen2-VL-7B and InternVL3-8B, Q-Instruct tuning actually results in performance degradation compared to their baseline versions. This unexpected finding suggests that stronger baselines may already possess strong general perception capabilities that can be disrupted if the training dataset lacks valuable IQA expertise.
3) The Qwen2-VL-7B model demonstrates exceptional performance when tuned with ViDA-UGC, achieving the highest overall score (80.6%) among all tested models, even surpassing GPT-4o's zero-shot performance (78.6%). This highlights the potential of targeted fine-tuning to elevate mid-sized models to performance levels comparable with much larger, more resource-intensive models.
4) Across all models, performance on "Yes-or-No" questions is consistently higher than on "What" and "How" questions, indicating that binary classification tasks remain easier than open-ended perception tasks requiring more detailed analysis.

Performance on ViDA-UGC-Bench is in Table 9. The results on ViDA-UGC-Bench reveal complementary insights:
1) Models exhibit varying capabilities across different distortion dimensions. "Clarity" and "Noise" dimensions generally see higher performance across models compared to "Compression" and "Content" dimensions, suggesting that certain types of distortions are inherently more challenging to assess accurately.
2) The impact of training datasets is even more pronounced on ViDA-UGC-Bench. Models tuned with ViDA-UGC consistently outperform both their baseline versions and Q-Instruct-tuned counterparts across all dimensions and concerns, often by substantial margins. This demonstrates the effectiveness of domain-specific training for UGC image quality assessment tasks.
3) Among assessment concerns, "Type" and "Position" generally receive higher scores than "Severity" and "Significance" across most models. This pattern indicates that identifying the category and location of distortions is easier than judging their impact and importance, which requires more nuanced perceptual reasoning.
4) While GPT-4o demonstrates strong zero-shot perfor-

Table 13: Comparison of the **low-level Perception** ability between baseline MLLMs, Q-Instruct-*tuned* versions, and **ViDA-UGC**-*tuned* versions on Q-Bench (LLVisionQA-*dev*). For each baseline model, the highest score is highlighted in **bold**.

| **Model** (*variant*) | Training Dataset | Yes-or-No↑ | What↑ | How↑ | Distortion↑ | Other↑ | I-C Distortion↑ | I-C Other↑ | Overall↑ |
|---|---|---|---|---|---|---|---|---|---|
| Qwen-VL-Chat | *no* (Baseline) | 56.00% | 58.63% | 54.77% | 50.78% | 61.57% | 53.62% | 62.45% | 56.39% |
| | Q-Instruct | **78.36%** | 74.56% | 61.05% | 70.43% | **67.59%** | 74.34% | **77.14%** | 71.51% |
| | **ViDA-UGC** | 77.09% | **78.32%** | **66.53%** | **78.60%** | 65.97% | **79.93%** | 71.02% | **73.98%** |
| Qwen2-VL-7B | *no* (Baseline) | 83.82% | 82.74% | 64.71% | 75.49% | **77.08%** | 75.66% | **82.86%** | 77.19% |
| | Q-Instruct | **84.00%** | 80.97% | 65.31% | 75.68% | 75.69% | 77.63% | 80.82% | 76.92% |
| | **ViDA-UGC** | 82.55% | **86.95%** | **72.62%** | **87.16%** | 70.83% | **85.20%** | 78.37% | **80.6%** |
| InternVL2.5-8B | *no* (Baseline) | 78.91% | 77.21% | 65.52% | 69.07% | **76.85%** | 70.07% | **84.08%** | 73.98% |
| | Q-Instruct | **81.64%** | 83.63% | 64.10% | 76.65% | 72.00% | 76.64% | 83.67% | 76.45% |
| | **ViDA-UGC** | 81.45% | **85.18%** | **71.60%** | **87.74%** | 66.20% | **84.54%** | 78.37% | **79.33%** |
| InternVL3-8B | *no* (Baseline) | 78.91% | 76.99% | **67.95%** | 70.43% | **75.93%** | 71.38% | **85.71%** | 74.72% |
| | Q-Instruct | 76.73% | 79.20% | 63.69% | 70.43% | 70.37% | 75.99% | 80.41% | 73.18% |
| | **ViDA-UGC** | **80.00%** | **85.40%** | 66.53% | **82.49%** | 70.37% | **79.93%** | 74.69% | **77.19%** |
| GPT-4o | *no* (Zero-shot) | 83.59% | 82.40% | 71.81% | 75.14% | 78.76% | 78.10% | 85.00% | 78.60% |

Table 14: Comparison of the **low-level perception** ability between baseline MLLMs, Q-Instruct-*tuned* versions, and **ViDA-UGC**-*tuned* versions on ViDA-UGC-Bench. For each baseline model, the highest score is highlighted in **bold**. Compress means compression dimension, and content means content anomaly dimension.

| **Model** (*variant*) | Training Dataset | Dimension | | | | | Concern | | | | Overall |
|---|---|---|---|---|---|---|---|---|---|---|---|
| | | *Clarity* | *Compress* | *Exposure* | *Content* | *Noise* | *Type* | *Position* | *Severity* | *Significance* | |
| Qwen-VL-Chat | *no* (Baseline) | 41.54% | 18.71% | 46.20% | 50.85% | 13.03% | 31.91% | 37.83% | 31.72% | 36.64% | 34.84% |
| | Q-Instruct | 30.27% | 39.94% | 50.22% | 47.46% | 14.23% | 28.50% | 35.79% | 31.72% | 47.12% | 35.88% |
| | **ViDA-UGC** | **66.89%** | **69.34%** | **58.71%** | **66.10%** | **46.86%** | **60.27%** | **54.44%** | **74.37%** | **69.49%** | **63.34%** |
| Qwen2-VL-7B | *no* (Baseline) | 51.90% | 52.83% | 43.75% | 47.46% | 17.15% | 43.43% | 43.53% | 51.26% | 54.15% | 47.53% |
| | Q-Instruct | 43.17% | 55.82% | 39.06% | 45.76% | 20.92% | 34.12% | 42.00% | 51.47% | 52.08% | 44.14% |
| | **ViDA-UGC** | **69.45%** | **84.12%** | **69.64%** | **86.44%** | **46.86%** | **76.37%** | **61.17%** | **80.46%** | **72.20%** | **71.45%** |
| InternVL2.5-8B | *no* (Baseline) | 39.75% | 53.77% | 44.64% | 55.93% | 18.41% | 37.08% | 43.78% | 44.96% | 49.04% | 43.51% |
| | Q-Instruct | 34.25% | 59.75% | 41.96% | 61.86% | 32.22% | 29.99% | 38.32% | 68.07% | 45.53% | 43.40% |
| | **ViDA-UGC** | **73.15%** | **85.69%** | **71.43%** | **84.75%** | **60.25%** | **81.24%** | **62.06%** | **82.14%** | **78.27%** | **74.80%** |
| InternVL3-8B | *no* (Baseline) | 50.76% | 52.52% | 46.65% | 55.93% | 12.13% | 43.57% | 47.08% | 47.06% | 52.08% | 47.37% |
| | Q-Instruct | 31.50% | 55.97% | 38.84% | 52.54% | 28.03% | 29.10% | 34.90% | 53.15% | 47.28% | 39.77% |
| | **ViDA-UGC** | **71.35%** | **87.74%** | **70.31%** | **83.90%** | **46.03%** | **82.87%** | **61.80%** | **78.15%** | **72.52%** | **73.00%** |
| GPT-4o | *no* (Zero-shot) | 54.93% | 51.26% | 72.54% | 58.47% | 26.36% | 46.38% | 60.28% | 53.57% | 59.58% | 55.20% |

mance, particularly in "Position" and "Type" concerns, it is outperformed by ViDA-UGC-tuned models in most categories. This suggests that while large foundation models possess impressive general capabilities, specialized training remains crucial for optimal performance in domain-specific tasks.

**Grounding Results.** For the distortion grounding task and region perception task, we train three object detection models (Feng et al. 2021; Zong, Song, and Liu 2023; Liu et al. 2024) with distortion bounding boxes and corresponding type. We compare ViDA-UGC-*tuned* MLLMs with detection models in Table 8. For the distortion grounding task, object detection models demonstrate superior performance to MLLM-based approaches. This is probably because they are better at the simple ten-class detection task. In MLLM-based approaches, the performance of generating quality descriptions interleaved with grounding information is comparable to that of directly outputting bounding box coordinates, highlighting the robustness of these approaches.

For region perception tasks, we crop the RoI sub-images and input them into detection models to obtain distortion bounding boxes. Table 8 showcases that MLLM-based approaches achieve better performance than the distortion grounding task, but still lag behind detection models. Nevertheless, MLLM-based methods demonstrate a significant advantage in versatility and capability over traditional detection methods, offering additional abilities such as answering IQA questions and integrating grounding into quality analysis. In summary, we maintain a comparable performance in the distortion grounding and region perception tasks.

## Qualitative Results

More qualitative results of low-level perception, quality description are presented in Figure 6 and Figure 7. Qwen2-VL-7B(ViDA-UGC) could accurately locate distortions, analyze their impacts on the display of image contents, then weigh the advantages and disadvantages of different aspects, and finally draw a final conclusion.

| Low clarity | Visually blurred, with both the "lines" and "textures" of the object being indistinct. | 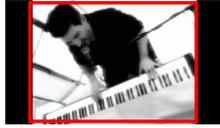 |
|---|---|---|
| Motion blur | Blurring occurs in images due to the rapid movement of the camera, the photographed object, or the scene during shooting. Objects in the image appear to have stretched trajectories and cannot be clearly distinguished. | 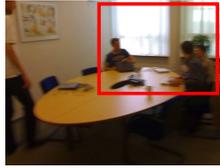 |
| Out of focus blur | The photographed object fails to be presented clearly with its details blurred due to improper focusing, which impairs the viewing experience. | 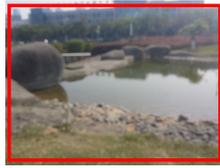 |
| Blocking artifacts | A type of distortion caused by image or video compression, characterized by the division of regions in the image into distinct blocks. | 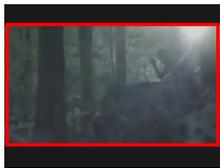 |
| Edge aliasing effect | A phenomenon where diagonal lines or edges in an image appear jagged, typically caused by insufficient resolution or an inadequate sampling rate, characterized by unsmooth edges. | 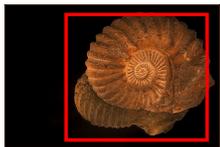 |
| Edge ringing effect | Oscillations appear at the edges of objects, similar to the air oscillations generated after a bell is struck. | 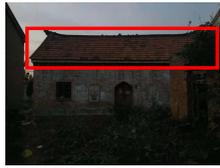 |
| Overexposure | Certain areas in the image become excessively bright due to overexposure, losing details and typically appearing as patches of white light. | 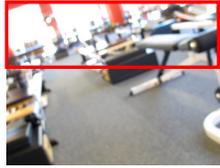 |
| Underexposure | Certain dark areas in the image become excessively dark due to underexposure, resulting in the loss of details and appearing as solid black regions. | 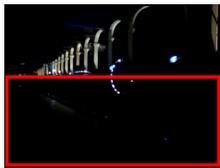 |
| Meaningless solid color | There are large solid-color areas in the picture, with text usually in the foreground. | 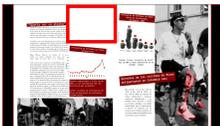 |
| Noise | Random bright or colored interfering spots appearing in images or videos, which are more noticeable under low-light conditions and typically appear as white (white noise) or colored (color noise). | 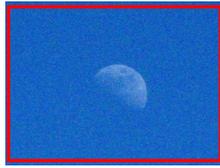 |

Figure 5: The definition and example images for each distortion type. The position of distortions are annotated by a red bounding box.

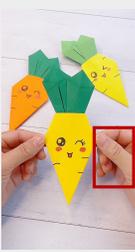

Figure 6: Qualitative results on low-level perception tasks. Corresponding distortions are annotated by the red bounding boxes.

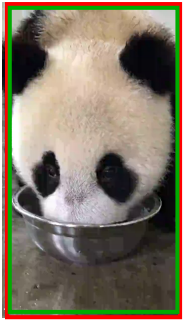

The image shows a close-up of a panda's face, with the panda holding a small bowl, possibly eating or playing with it.
There is one **moderate blocking artifacts(1,3),(992,998)** affecting the panda's face and the bowl. The blocking artifacts manifest as square-like patterns that disrupt the smoothness of the panda's fur and the bowl's surface, creating a pixelated effect. This reduces the clarity and sharpness of the panda's fur and facial features, making them appear less detailed and slightly pixelated. The texture of the fur and the bowl is compromised, affecting the overall visual quality.
There is one **moderate low clarity(18,6),(989,999)** affecting the panda's face and the bowl. The low clarity distortion causes the panda's face and the bowl to appear slightly blurred, with less defined edges and a softer appearance. This reduces the sharpness and detail of the panda's facial features and the bowl, making them appear less distinct and slightly blurred.
Both distortions affect the essential region of the panda's face and the bowl, which are central to the image's focus. The blocking artifacts introduce a pixelated effect that disrupts the natural texture, while the low clarity further blurs the details, resulting in a lack of sharpness and definition. These combined effects significantly reduce the visual appeal and detail of the image, making it difficult to appreciate the subject's features. The image's low-level attributes include a lack of sharpness and detail, with a pixelated texture that detracts from the overall visual experience.
Thus, the quality of the image is poor.

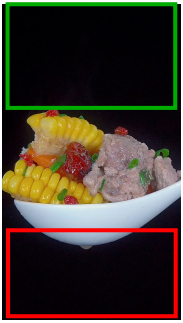

The image shows a close-up of a spoonful of food, featuring corn, meat, green vegetables, and possibly goji berries, set against a dark background.
There are two occurrences of meaningless solid color distortion. **Meaningless solid color(15,11),(983,298)** appears in the dark background above the spoon, creating a uniform area that lacks texture or detail. This reduces the contrast between the spoon and the background, potentially affecting the overall composition. Similarly, **meaningless solid color(22,723),(997,979)** is present in the dark background below the spoon, resulting in a flat, featureless area that diminishes the visual separation between the spoon and the background.
These meaningless solid colo distortions create areas of uniformity that reduce the contrast and visual separation between the spoon and the background, impacting the overall composition. Despite these issues, the image maintains a good level of detail and color contrast in the spoon's contents, which helps preserve some visual interest. The low-level attributes of the image include a lack of texture in the background and a generally dark tone, which affects the overall perception of depth and detail.
Thus, the quality of the image is fair.

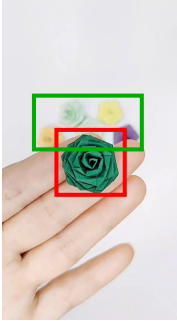

The image shows a hand holding a small, intricately folded green origami rose, with several other origami roses in different colors placed on a white surface in the background.
There is one **moderate out of focus blur(169,277),(818,416)** affecting the origami roses in the background. The origami roses appear blurred and lack detail, with their edges and folds being indistinct. This affects the visibility and sharpness of the roses, making it difficult to appreciate their details and colors.
There is one **moderate low clarity(311,403),(657,619)** affecting the green origami rose held in the hand. The edges and details of the rose appear less defined, with the folds of the paper being somewhat blurred. This reduces the crispness of the rose's appearance and diminishes the perception of texture and precision.
The **moderate low clarity(311,403),(657,619)** affecting the green origami rose held in the hand has the most significant impact on the overall quality. As the focal point, the lack of sharpness and definition in the rose's details detracts from the intended visual appeal of the composition. The image's low-level attributes include a lack of distinct edges and a generally soft appearance, with no significant color distortions present.
Thus, the quality of the image is fair.

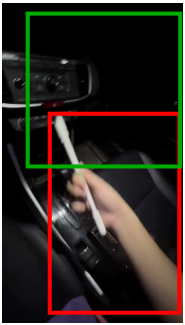

The image shows the interior of a car, focusing on the center console and dashboard, with a hand holding a white cleaning brush.
There is one motion blur in the image. **Moderate motion blur(271,330),(972,987)** affects the hand holding the cleaning brush and the area of the center console it is cleaning. The hand and cleaning brush appear with stretched trajectories, leading to a loss of detail and sharpness. The motion blur creates an elongated effect on the objects, making them appear smeared across the image.
There are two underexposures in the image. **Moderate underexposure(372,152),(999,520)** affects the dashboard controls and buttons in the upper area. The affected area appears darker than the rest of the image, with reduced visibility of the controls and buttons.
The **moderate motion blur(271,330),(972,987)** of the hand and cleaning brush is the most significant distortion, as this region is essential for understanding the primary action in the image. The blurring severely impacts the viewer's ability to discern the action being performed. Additionally, **moderate underexposure(372,152),(999,520)** in the upper-left dashboard area also plays a crucial role, as it affects an important part of the image, further diminishing the visibility and clarity of essential controls. The overall image appears dark, with significant areas lacking detail due to underexposure, and the motion blur contributes to a lack of sharpness and definition across critical regions. These issues collectively result in a diminished perception of the car's interior quality and a reduced ability to engage with the image.
Thus, the quality of the image is poor.

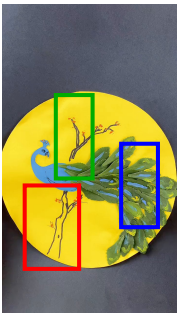

The image features a circular artwork with a yellow background, showcasing a blue peacock with a body painted and a tail creatively made using green leaves, perched on a branch surrounded by additional branches with small orange flowers.
There are three edge aliasing effects in the image. **Moderate edge aliasing effect(531,503),(892,749)** affects the peacock's tail made of green leaves, causing the edges of the leaves to appear jagged and uneven, disrupting their natural shape. **Moderate edge aliasing effect(217,566),(458,835)** affects the branch the peacock is perched on, causing the edges to appear jagged and pixelated, impacting the clarity and detail of the branch. **Moderate edge aliasing effect(388,395),(562,499)** affects the small branches with orange flowers in the upper center, causing the edges to appear jagged and pixelated, impacting their visibility and sharpness.
The **moderate edge aliasing effect(217,566),(458,835)** on the branch the peacock is perched on has the most significant impact on the overall image quality due to its essential role in the composition. The jagged and pixelated appearance of the branch reduces its clarity and sharpness, affecting the viewer's perception of the main subject. The low-level attributes of the image include a vibrant color palette dominated by yellow, blue, and green, which enhances the visual appeal despite the distortions. The artwork's composition relies heavily on the central peacock and its integration with surrounding elements, making the clarity of these components crucial for the visual harmony of the image. Thus, the quality of the image is fair."

Figure 7: Qualitative results on quality description tasks.

 
\clearpage
\end{document}